\documentclass[final]{cvpr}

\usepackage{times}
\usepackage{epsfig}
\usepackage{graphicx}
\usepackage{amsmath}
\usepackage{amssymb}
\usepackage[caption=false, font=footnotesize]{subfig}
\usepackage{nopageno}

\usepackage{pifont}% http://ctan.org/pkg/pifont
\newcommand{\cmark}{\ding{51}}%
\newcommand{\xmark}{\ding{55}}%

\newcommand\blfootnote[1]{%
  \begingroup
  \renewcommand\thefootnote{}\footnote{#1}%
  \addtocounter{footnote}{-1}%
  \endgroup
}

\usepackage[pagebackref=true,breaklinks=true,colorlinks,bookmarks=false]{hyperref}
\usepackage{multicol}

\def\cvprPaperID{33} % *** Enter the CVPR Paper ID here
\def\confYear{AI4Space 2021}
\begin{document}

%%%%%%%%% TITLE
\title{LSPnet: A 2D Localization-oriented Spacecraft Pose Estimation Neural Network}

\author{Albert Garcia$^{1}$%\\{\tt\small albert.garcia@uni.lu}
% For a paper whose authors are all at the same institution,
% omit the following lines up until the closing ``}''.
% Additional authors and addresses can be added with ``\and'',
% just like the second author.
% To save space, use either the email address or home page, not both
% \and
% Second Author\\
% Institution2\\
% First line of institution2 address\\
% {\tt\small secondauthor@i2.org}
\and
Mohamed Adel Musallam$^{1}$%\\{\tt\small mohamed.ali@uni.lu}
\and
Vincent Gaudilliere$^{1}$%\\{\tt\small vincent.gaudilliere@uni.lu}
\and
Enjie Ghorbel$^{1}$%\\{\tt\small enjie.ghorbel@uni.lu}
\and
Kassem Al Ismaeil$^{1}$%\\{\tt\small kassem.alismaeil@uni.lu}
\and
Marcos Perez$^{2}$%\\{\tt\small m.perez@lmo.space}
\and
Djamila Aouada$^{1}$%\\{\tt\small djamila.aouada@uni.lu}
\and}

\affiliation{
$^{1}$ Interdisciplinary Center for Security, Reliability and Trust (SnT)\\University of Luxembourg, Luxembourg\\
{\tt\small \{albert.garcia, mohamed.ali, vincent.gaudilliere, enjie.ghorbel}\\
{\tt\small kassem.alismaeil, djamila.aouada\}@uni.lu}\\
% {\tt\small firstname.lastname@uni.lu}
\\
$^{2}$ LMO\\
{\tt\small m.perez@lmo.space}
}

\maketitle

%%%%%%%%% ABSTRACT
\begin{abstract}
Being capable of estimating the pose of uncooperative objects in space has been proposed as a key asset for enabling safe close-proximity operations such as space rendezvous, in-orbit servicing and active debris removal. Usual approaches for pose estimation involve classical computer vision-based solutions or the application of Deep Learning (DL) techniques. This work explores a novel DL-based methodology, using Convolutional Neural Networks (CNNs), for estimating the pose of uncooperative spacecrafts. Contrary to other approaches, the proposed CNN directly regresses poses without needing any prior 3D information. Moreover, bounding boxes of the spacecraft in the image are predicted in a simple, yet efficient manner.
% The proposed technique differs from the rest of the space literature by not relying on classical feature descriptors as well as not implementing the popularly used object detection models from DL literature. Conversely, the here proposed work offers an alternative approach to object detection through a 2D localization-oriented methodology which yields bounding boxes in a direct manner.
The performed experiments show how this work competes with the state-of-the-art in uncooperative spacecraft pose estimation, including works which require 3D information as well as works which predict bounding boxes through sophisticated CNNs.
\blfootnote{This  work  was  funded  by  the  Luxembourg  National  Research  Fund (FNR),  under  the  project  reference BRIDGES2020/IS/14755859/MEET-A/Aouada, and by LMO (https://www.lmo.space).}
\end{abstract}

%%%%%%%%% BODY TEXT
\section{Introduction}
In recent years, more and more space mission scenarios have involved close-proximity operations with uncooperative space objects such as space debris (\eg active debris removal), out-of-order satellites (in-orbit servicing) or even comets and asteroids (space exploration).\\
% Closely related to this, there has been an impactful increase in the number of space technology start-ups which aim to address the aforementioned close-proximity operations. Some examples of such start-ups are LMO\footnote{https://www.lmo.space/}, ClearSpace\footnote{https://clearspace.today/} and Infinite Orbits\footnote{https://www.infiniteorbits.io/}.
\begin{figure}[t]
\begin{center}
   \includegraphics[width=0.95\linewidth]{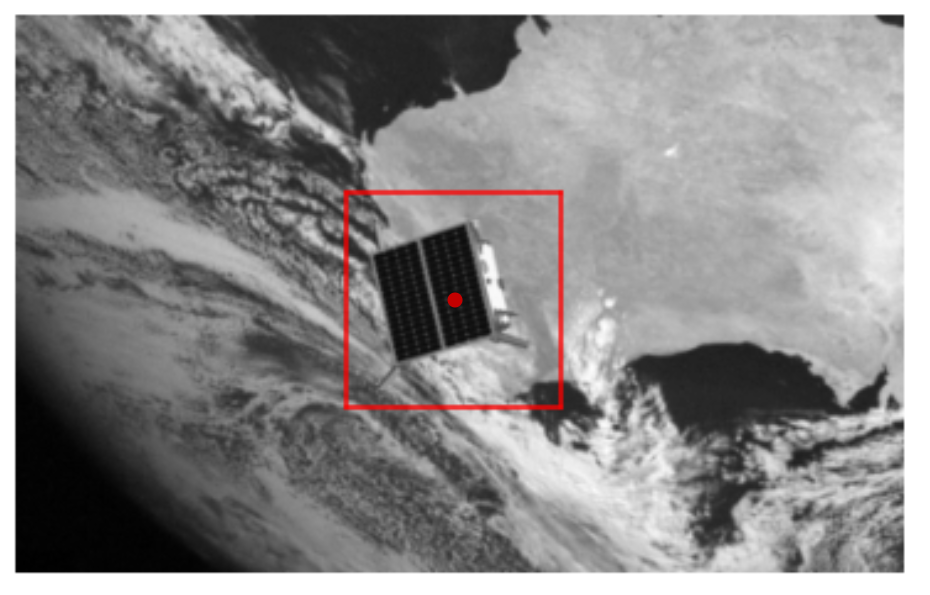}
    \end{center}
   \caption{Example of a spacecraft bounding box detection by LSPnet. The bounding box is constructed based on the central red dot yielded by LSPnet.}
\label{fig:Input-image-ROI-red-dot}
\end{figure}
In these scenarios, a chaser spacecraft seeks to approach then to capture or to dock at a target orbiting space object, that is uncooperative. In other words, the latter is not communicating any information to the former, either actively (\eg by radio communication) or passively (\eg by featuring a fiducial marker)~\cite{OPROMOLLA201753}. In this context, most space relative navigation methods first require to estimate the relative position and attitude, referred to as pose, between both spacecrafts, then to track the relative pose over time using previous estimations~\cite{review}. To enable autonomous close-proximity operations in space with uncooperative targets, robust and efficient on-board pose initialization solutions are required. To this end, several vision-based works propose to use active sensors such as Light Detection and Ranging (LIDAR)~\cite{OPROMOLLA2015287, SPARK}. Despite their demonstrated efficiency, such sensors remain heavy and high-power consuming. On the contrary, relying on a single monocular camera has the advantage of complying with strict power and mass requirements relating to space missions, while ensuring a low level of system complexity~\cite{review}. However, computing the pose of a known uncooperative object from a single monocular camera is a challenging task. First of all, the target object in the field of view of the camera can be depicted in a wide range of different scales, depending on the target's size and on the distance to the chaser spacecraft. Secondly, in a large number of cases, it is necessary to deal with cluttered backgrounds introduced by the Earth which can heavily complicate the task of pose estimation. Finally, due to the nature of the input data, a target detection step over the captured image is desired to reliably perform orientation estimation as in the latest Satellite Pose Estimation Challenge (SPEC)~\cite{SPEC}. The downside of including a detection step is the increase in the solution complexity as well as the decrease of its computational efficiency.

\begin{figure}[t]
\begin{center}
   \includegraphics[width=0.95\linewidth]{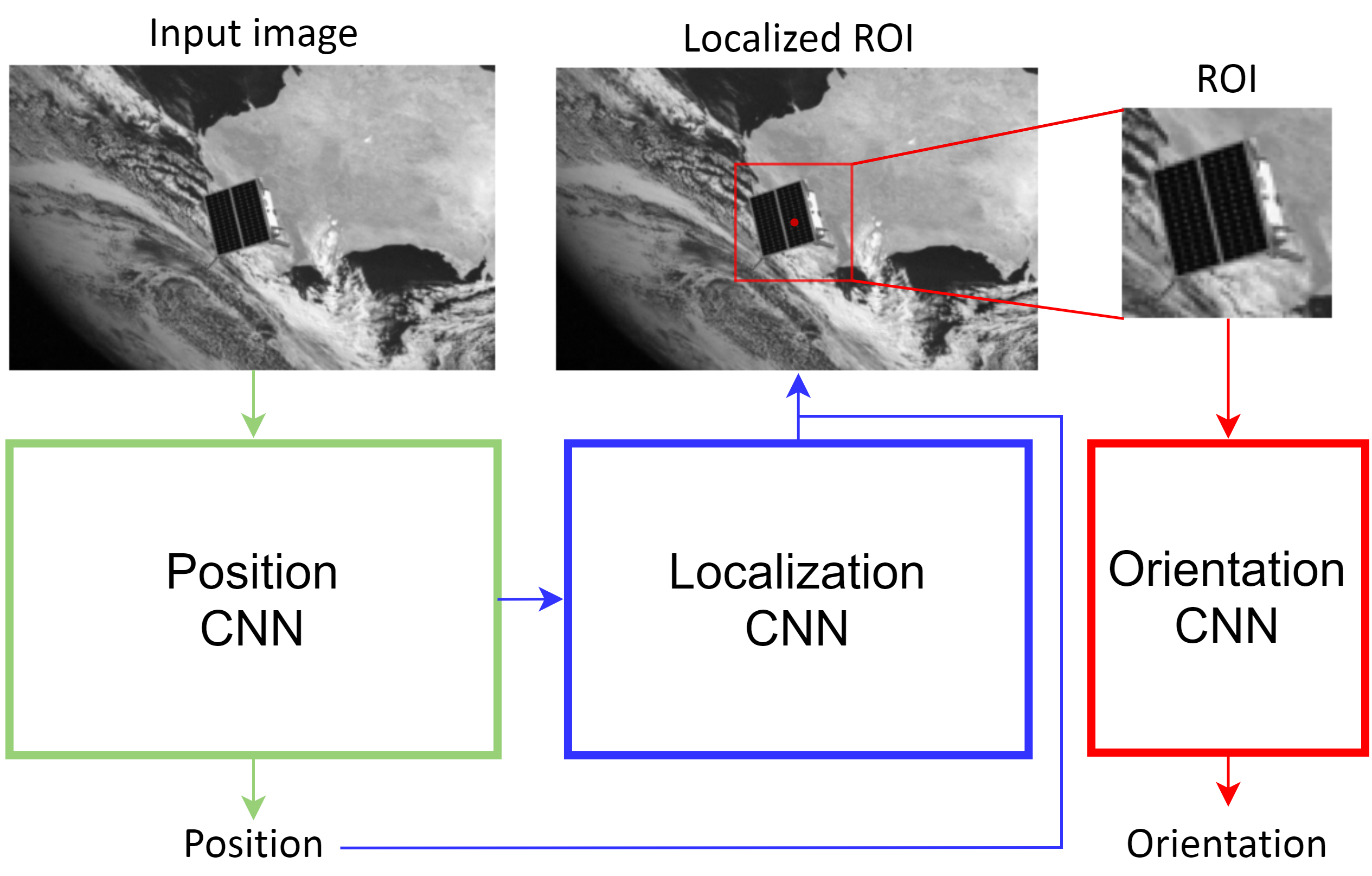}
    \end{center}
   \caption{High-level architecture overview of the proposed LSPnet. Given an input image with a target spacecraft, LSPnet yields its position and orientation as well as a bounding box. Highlighted in green are the parts of LSPnet aimed towards position estimation, in red the ones aimed towards orientation estimation while in blue are the parts which support both tasks.}
\label{fig:LSPnet-overview}
\end{figure}

The solution proposed herein, named \emph{2D Localization-oriented Spacecraft Pose Estimation Neural Network} (LSPnet), deals with the aforementioned challenges while remaining simple and efficient. Our work takes advantage of the Deep Learning (DL)-based advances in Computer Vision by implementing a Convolutional Neural Network (CNN). In contrast to the common approach of implementing an object detection network (such as YOLO~\cite{YOLO} or Faster-RCNN \cite{Faster-RCNN}) for the detection step, our approach is capable of yielding a simple bounding box in a straightforward manner. Additionally, a 2D-localization process is developed in order to aid the part of the network responsible for 3D position estimation. In other words, LSPnet learns how to estimate the spacecraft position while being driven by an auxiliary network which detects the center of the spacecraft in the image. Finally, combining the predicted position into the center detection network, a region of interest (ROI) crop (see Figure~\ref{fig:Input-image-ROI-red-dot}) is performed and used as input for orientation estimation, thus yielding the full pose of the uncooperative target spacecraft. Figure~\ref{fig:LSPnet-overview} presents a high-level overview of LSPnet as well as the connections between its modules.

The remainder of the paper is organized as follows: Section~2 reviews the related literature on spacecraft pose estimation. Section~3 formulates the problem and the proposed approach. Section~4 describes and discusses the experimental evaluation. Lastly, Section~5 concludes the paper.

%-------------------------------------------------------------------------

\section{Related work}

Spacecraft pose estimation from a single monocular camera has extensively drawn techniques from Computer Vision literature. Thus, we provide an overview of the different approaches that appear in space applications. %while presenting their state-of-the-art.

\subsection{Model-based approaches}
Many pose estimation methods rely on a 3D model of the target spacecraft. One of the most proposed approaches consists in matching the 2D input image with a 3D wireframe of the spacecraft. To do so, visual features are extracted from the image then matched to their corresponding elements in the wireframe. In these cases, the final pose is obtained by solving the Perspective-n-Point (PnP) problem. The features used to solve PnP vary between interest points (keypoints), corners, edges and depth maps. Classical works use handcrafted Computer Vision filters to detect these features such as Canny and Sobel filters~\cite{sobel,canny}. State-of-the-art in handcrafted Computer Vision filtering extracts a range of different features which are afterwards fused into a dense feature representation~\cite{sota-handcrafted}.

In addition to keypoint-based solutions which aim to predict the pose by performing 2D-3D matching, there are other approaches that also rely on  the spacecraft 3D model. Another common model-based approach is to minimize the \textit{projection} error defined as the misalignment between the spacecraft in the image and the projected 3D model by the predicted pose. These approaches require an initial pose which is afterwards refined through a \textit{projection} error minimization process. For instance, the work of \cite{projection-paper} first initializes a rough pose by means of feature matching and then refines it through a multi-dimensional Newton-Raphson algorithm used to minimize a projection error. Similarly, the work of \cite{sharma2017reduced} fuses a weak gradient elimination technique to detect finer features and estimates the pose based on a Newton-Raphson projection minimization fashion.

% Another common model-based approach is to minimize the \textit{projection} error defined as the misalignment between the spacecraft in the image and the projected 3D model by the predicted pose. \cite{projection-paper} first initializes a rough pose and then refines it through an iterative optimization procedure involving the projection of a simplified 3D model. Related to this, it is common practice to estimate an initial pose which is further refined through an iterative process. For instance, the work of \cite{SPEC-Rank1} initializes the pose through PnP by its predicted 2D-3D correspondences and afterwards refines the pose following a non-linear optimization process.

\subsection{Appearance-based approaches}
In comparison to feature-based methods, some approaches rely on directly exploiting the \textit{appearance} of the spacecraft in the image. To the best of our knowledge, the only appearance-based method using a monocular camera for spacecraft pose estimation is the work of \cite{appearance-based}. This method performs Principal Component Analysis (PCA) over the spacecraft present in a query image in order to match it to a dataset of stored images with their corresponding pose ground truths. By performing PCA, they drastically reduce the dimensions of the dataset. Despite this, the proposed method still requires to compare the query image to each entry of the stored dataset, thus making it not scalable \cite{review}.

\subsection{Deep Learning-based approaches}
In recent years, there has been a clear trend to rely on DL techniques in order to perform spacecraft pose estimation. The latest SPEC challenge~\cite{SPEC} informs of a clear dominance in DL-based solutions among the participant teams. Following this trend, several works aim to directly regress the pose of the spacecraft through CNNs such as the Spacecraft Pose Network (SPN) presented in \cite{SPEED}, the network proposed in \cite{SPEC-Rank3} or the off-the-shelf GoogLeNet CNN~\cite{GoogLeNet} implemented in \cite{phisannupawong2020vision}. The recent work of \cite{sonawani2020assistive} implements a double VGG architecture \cite{VGG} to directly regress translation and rotation over a synthetic dataset as well as over a laboratory-acquired dataset simulating an on-orbit assembly operation.
Deep Learning techniques offer great robustness against different lighting scenarios as well as robustness against cluttered backgrounds \cite{Learning-methods-Light-Background}.
Other works combine Deep Learning with classical approaches, \eg Deep Learning keypoints regression combined with PnP solving. The works of \cite{SPEC-Rank1, SPEC-baseline, black2021real} all perform a first step of zooming in into a ROI yielded by an Object detection neural network. Afterwards, \cite{SPEC-Rank1, black2021real} regress a set of manually selected keypoints while \cite{SPEC-baseline} regresses the corners of the target spacecraft in an ordered manner to avoid additional matching computations.
% \cite{huo2020fast} fuses together the Object detection task with keypoints regression in a unique CNN while achieving remarkable results at low processing times.
Keypoint-based pose estimation solutions are generally robust and accurate provided that high quality 2D-3D correspondences can be obtained beforehand. Variations in lighting conditions as well as occluded keypoints can heavily impact pose accuracy. Fortunately, Deep Learning-based techniques have proven to efficiently handle these scenarios thanks to their generalization capabilities \cite{Learning-methods-Light-Background}.

% Recently, due to the huge success of Deep Learning applications, modern Computer Vision techniques involve Deep Learning neural networks in order to perform feature detection. \cite{SPEC-baseline} takes a similar approach but focuses on predicting corners of the target spacecraft in an ordered-manner to avoid additional matching computations. Keypoint-based pose estimation solutions are generally robust and accurate provided that high quality 2D-3D correspondences can be obtained beforehand. Variations in lighting conditions as well as occluded keypoints can heavily impact pose accuracy.

% Survey \cite{cassinis2019review} collects and categorizes an extensive list of monocular-based solutions for uncooperative spacecraft pose estimation. We let the reader refer to \cite{cassinis2019review} for further contextualization.

%-------------------------------------------------------------------------

\section{Proposed approach}
Formally, the problem statement of this work is the prediction of the object's pose, \textit{i.e.} the pose of \textit{O}, relative to the camera frame \textit{C}. In other words, the goal of the presented scenario is to predict the origin of the object's reference frame as well as its axes with respect to the camera's reference frame. This goal is achieved by estimating both the translation vector $t=(x,y,z)$ and the rotation matrix $R$ which transforms the reference frame of \textit{C} into the reference frame of \textit{O}. Both $t$ and $R$ are expressed in the camera basis meaning that the $z$ coordinate of the translation vector expresses the distance to the object \textit{O}. The rotation $R$, which is expressed as a $3\times3$ orthogonal real matrix, can also be represented by a quaternion $q=(q_1, q_2, q_3, q_4)$ with unit norm. In doing so, several advantages appear such as eliminating the gimbal lock problem as well as encoding the same rotation using only $4$ values instead of $9$. The quaternion $q$ encodes the same rotation through a closed-form mathematical formulation of a rotation axis and the angle to apply around the rotation axis. Given an input image $I$ which depicts a target spacecraft obtained through a monocular visual sensor, composed by a single channel (gray) or by three channels (red, green and blue), and given the here designed Deep Learning network LSPnet with optimized weights $w$, then $LSPnet(I,w)=(t, q)$.

LSPnet is formed by three interconnected CNNs (named Position, Localization and Orientation) as depicted in Figure \ref{fig:LSPnet-overview}. When grouping together the Position and Localization CNNs they can be referred to as the Translation module due to their collaborative behavior for optimizing the predicted translation $t$. Once the Orientation CNN is connected to the Translation module, a full Pose estimation module is formed. The following sections cover the Translation module, the complete Pose module, a connection variation between the two modules and a specifically designed data augmentation technique.

\begin{figure}[t]
      \centering
      \includegraphics[width=0.45\textwidth]{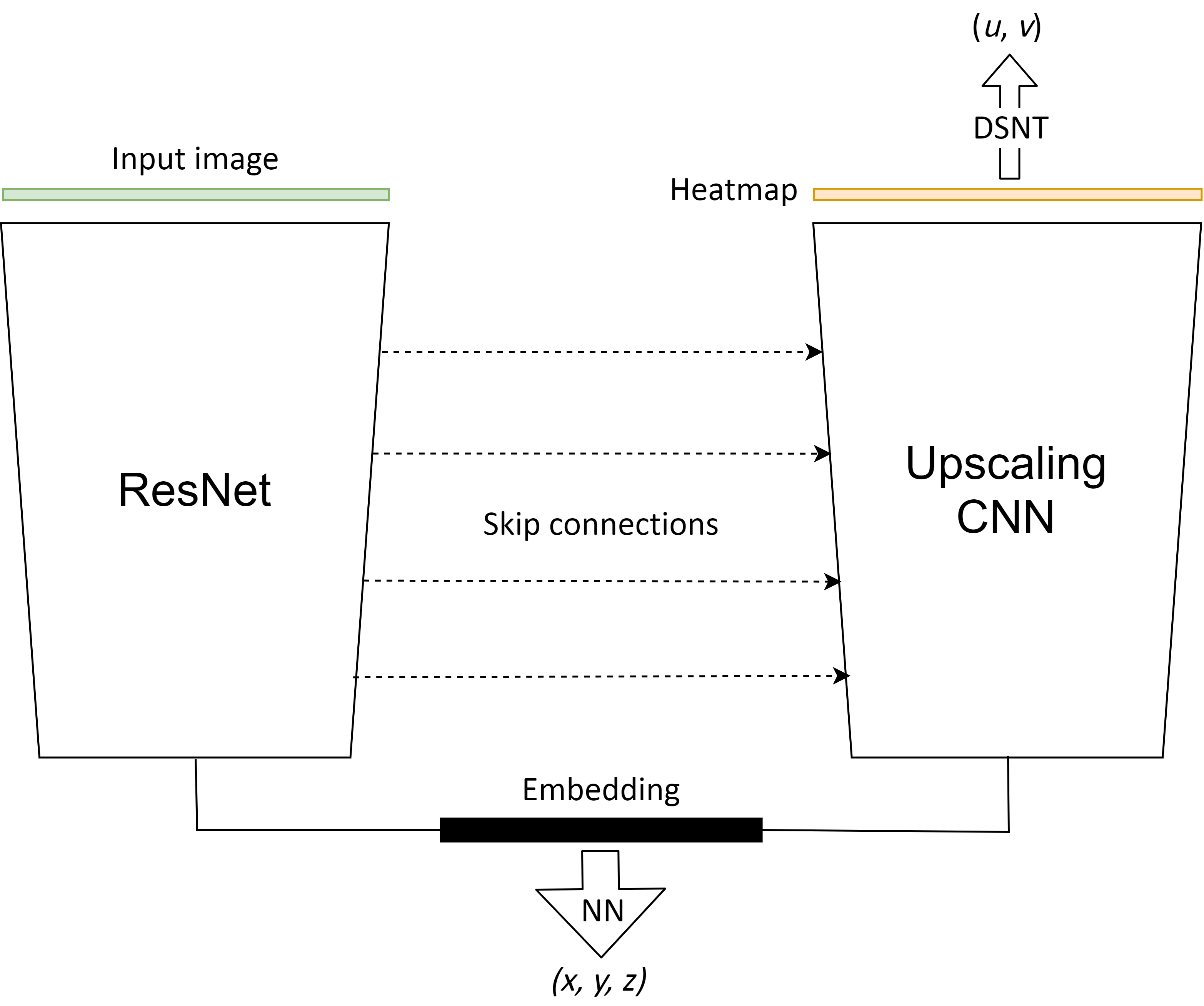}
      \caption{Architecture overview of the Translation module for the prediction of the 3-dimensional translation vector, $t=(x,y,z)$, as well as for the prediction of the pixel coordinates center of $O$, $(u,v)$.}
      \label{fig:TranslModule-overview}
\end{figure}
\subsection{Translation module}
Based on the successful CNN architecture named Unet \cite{Unet}, popularly used for image segmentation, as well as based on the scalable ResNet \cite{ResNet} architecture, famous for its breakthrough success in image recognition, we combine both methodologies into a unified CNN which composes the Translation module. We cast the goal of the Unet architecture, \textit{i.e.} image segmentation, into pixel coordinates regression through the inclusion of the transformation titled \textit{Differentiable spatial to numerical transform} (DSNT) \cite{Dsnt}. DSNT is designed to be a non-trainable Neural Network layer for transforming heatmaps into 2-dimensional coordinates in a differentiable manner, thus avoiding a direct decoupling of the predicted heatmaps from the loss function to be optimized. Combining the 2D spatial information conservation capability of Unet, the scalability of ResNet and the relevance in coordinates regression of DSNT, the final Translation module is formed. Figure \ref{fig:TranslModule-overview} offers an overview of the here described architecture for regressing the translation $t=(x,y,z)$ as well as the pixel coordinates $(u,v)$ which localizes the position of the object $O$ in the image. For practical reasons, the ground truth $(u,v)$ values can be derived from projecting the ground truth coordinates $x$ and $y$ from the translation vector into the image plane. In this case the pixel coordinates $(u,v)$ represent the center of the object's reference frame in the image pixel space.

Given the input image $I$, ResNet extracts an embedding which encodes 2D spatial information related to the target spacecraft. The extracted embedding is transferred through two different paths. First of all, a Neural Network (NN) formed by Fully Connected and ReLu layers is responsible for predicting the translation vector $t$. The second path connects to an upscaling CNN responsible for transforming the embedding back to the original size of the input $I$. This CNN makes use of upscaling layers followed by convolutional layers to increase the size of the embedding until reaching the original size. Additionally, following the Unet architecture, intermediate tensors from ResNet are concatenated into intermediate layers of the upscaling CNN. These connections are also known as skip connections. It is surmised that, through the use of these skip connections, different scales of the target spacecraft depicted in the image can be efficiently handled. The reason for this comes from the fact that the skip connections happen at different layers of ResNet, and thus the visual receptive field at each skip connection is different (from small to big scale features being captured). Once the embedding has been upscaled to the original size, it is convoluted to only present one channel and, afterwards, it is normalized. This normalized one-channel image corresponds to a heatmap representing the probabilities in the image for the pixel coordinates $(u,v)$ of the spacecraft's center. DSNT takes as input this normalized heatmap and regresses $(u,v)$. This same DSNT layer is the responsible for inducing the upscaling CNN towards predicting meaningful heatmaps. The upscaling CNN, which focuses on localizing the center of the spacecraft in the image pixel space, induces a localization-oriented training to ResNet when optimizing the embedding for translation estimation.

\subsection{Pose module}
The Pose module is built on top of the Translation module. Given the architecture as well as the outputs of the Translation module, the Pose module requires several components extracted from the former in order to estimate the quaternion $q$. First of all, the predicted pixel coordinates $(u,v)$ are taken and further processed in order to find a ROI and zoom over it. This ROI crop, yielded by a straightforward bounding box technique, is used for the estimation of the orientation through a ResNet CNN. The technique used for predicting the bounding box is as follows:
\begin{enumerate}
    \item The center of the bounding box in the image $I$ corresponds to the predicted pixel coordinates $(u,v)$.
    \item The bounding box always takes the shape of a square which should contain the spacecraft (on its entirety if possible).
    \item Knowing that the bounding box is a square, the only unknown variable left to estimate is its side length. Taking the predicted distance from the camera to the object, encoded in the $z$ component of the translation vector $t$ in $meters$ unit, a scaling transformation is applied as follows
    \begin{equation}
        BBL = \frac{K_{O}}{z}
    \end{equation}
    where $BBL$ is the length of the square bounding box in $pixels$ and $K_O$ is a constant parameter in $pixels \times meters$ unit which depends on the spacecraft being processed in the image.
\end{enumerate}
The hyperparameter $K_{O}$ needs to be fine-tuned depending on each spacecraft object $O$. This parameter encodes the size of the spacecraft in relation to the camera parameters. The predicted bounding box lets LSPnet zoom in into a ROI which has a significantly higher signal to noise ratio. Finally, the cropped ROI is then rescaled to a fixed size in order to process it through a ResNet CNN responsible for the regression of the orientation in quaternion form. Note how the proposed bounding box technique presents the advantage of not requiring any object detection ground truth labels. It is worth mentioning that after ResNet yields a 4-dimensional vector (in combination with Fully Connected and ReLu layers), it is normalized to impose the predicted quaternion to have unit norm. In addition to the here described methodology for estimating the orientation, a neural network data-flow connection can be added to the Orientation CNN in hopes of improving its accuracy.

\begin{figure*}[t]
      \centering
      \includegraphics[width=0.95\textwidth]{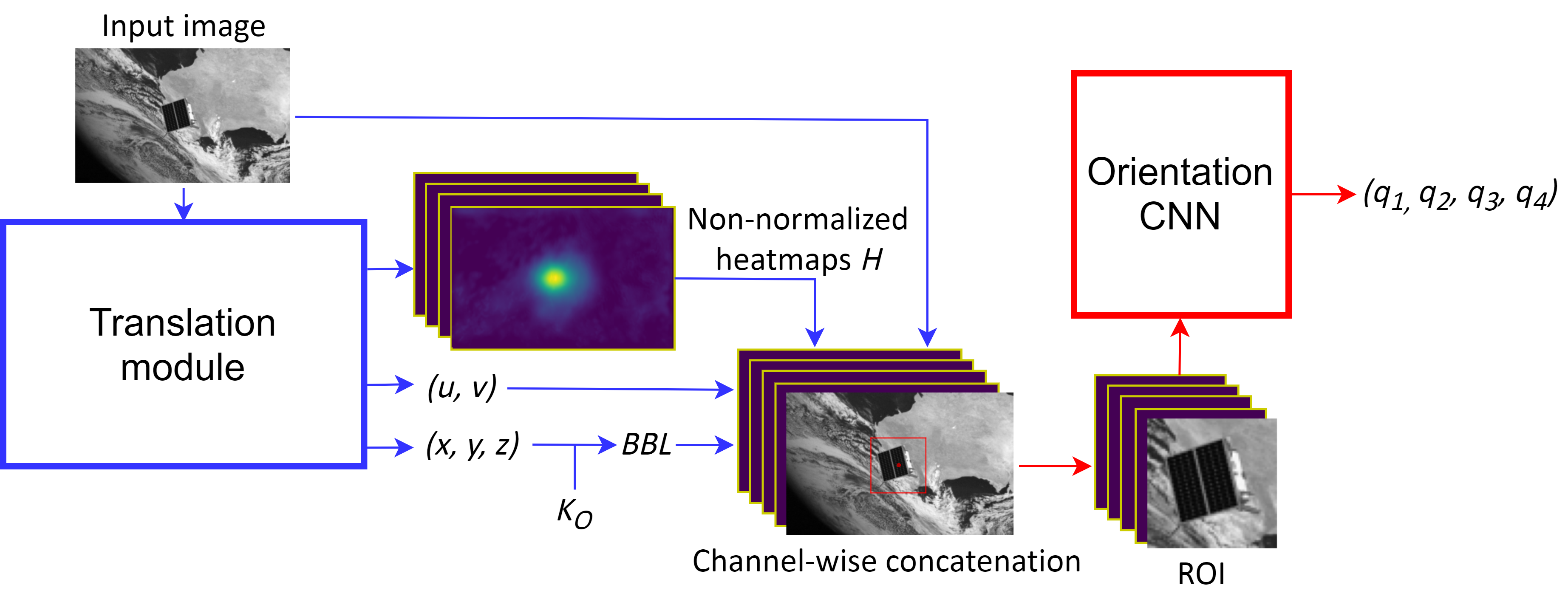}
      \caption{Diagram of the full Pose module pipeline which includes HC. For architectural details of the Translation module refer to Section~3.1 and to Figure \ref{fig:TranslModule-overview}.}
      \label{fig:LSPnet-concatenation}
\end{figure*}

\subsubsection{Heatmap Concatenation}
Prior to performing the ROI crop, \textit{i.e.} when taking the predicted bounding box and zooming on it over the input image $I$, the $H$ non-normalized heatmaps yielded by the Localization CNN (which are afterwards combined into a single normalized heatmap) can be concatenated channel-wise into the image $I$. For better clarification through the rest of the paper, we label this process as Heatmap Concatenation (HC). This technique offers the Orientation CNN an input tensor with $H + C_{I}$ channels where $C_{I}$ is the number of channels of $I$. By implementing HC we are creating a differentiable data-flow from the Orientation CNN into the whole Translation module. This in turn means that LSPnet can be entirely optimized at the same time in a fully-differentiable manner. An ablation study presented in Section~4.1. offers insights on the impact of including HC on LSPnet. Figure \ref{fig:LSPnet-concatenation} shows a diagram of the connections between the Translation module and the Orientation CNN in order to compose the final Pose module pipeline which includes HC.

\subsubsection{Center Data Augmentation}
\begin{figure*}[t]
      \centering
      \includegraphics[width=0.9\textwidth]{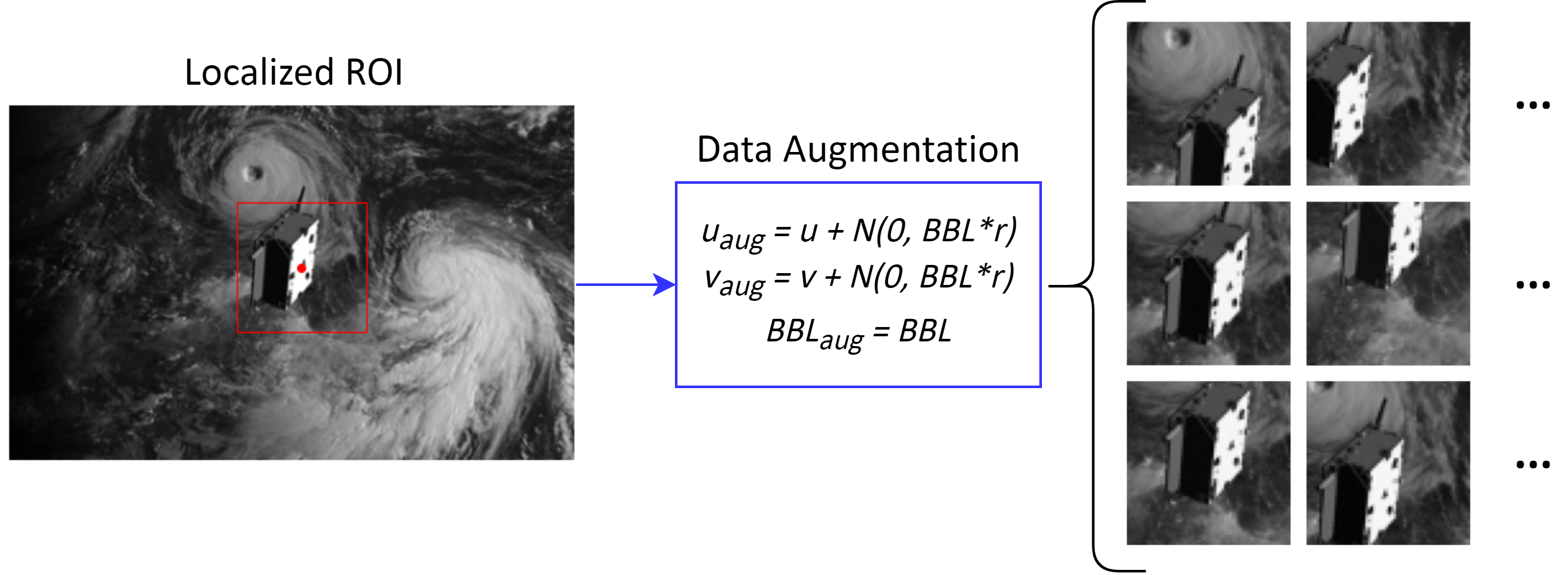}
      \caption{Data augmentation technique, named CDA, for augmenting the bounding boxes extracted from the outputs of the Translation module. The depicted example uses a value of $r=0.15$ meaning that the standard deviation of each Normal distribution equals to $15\%$ of the bounding box length.}
      \label{fig:LSPnet-center-data-augmentation}
\end{figure*}

Due to the nature of the bounding box methodology, a specialized data augmentation technique, which we name Center Data Augmentation (CDA), can be implemented in order to significantly increase the data variance offered to the Orientation CNN. The implemented data augmentation technique proposes a new bounding box based on the predicted one as follows
\begin{equation}
    u_{aug} = u + \mathcal{N}(0,\,BBL*r)
\end{equation}
\begin{equation}
    v_{aug} = v + \mathcal{N}(0,\,BBL*r)
\end{equation}
\begin{equation}
    BBL_{aug} = BBL
\end{equation}
where $(u_{aug}, v_{aug})$ is the center of the augmented bounding box, $BBL_{aug}$ is the augmented bounding box length and $r$ is a fixed hyperparameter such that $r \in \mathbb{R}$ and $r > 0$. The hyperparameter $r$ encodes the dispersion added to the center coordinates in relation to the length of the bounding box. To ensure that the augmented bounding box offers a high signal to noise ratio then small values of $r$ should be selected. An example of the results of CDA is depicted in Figure \ref{fig:LSPnet-center-data-augmentation}. It is worth remarking how CDA is able to provide challenging samples (due to truncated spacecrafts) as well as high signal to noise ratio samples as seen in Figure \ref{fig:LSPnet-center-data-augmentation}. To conclude, both HC and CDA are fully compatible and can be combined in hopes of further enhancing LSPnet.

%-------------------------------------------------------------------------

\section{Experiments}
A series of trainings and experimental setups have been carried out using the Spacecraft Pose Estimation Dataset (SPEED) \cite{SPEED}. This dataset offers a set of $12,000$ gray-scale pose-labeled synthetic images of size $1200 \times 1920$ portraying a spacecraft in space. The rendered images cover a wide range of different distances from the camera to the satellite (from $5m$ to $40m$ approximately). The SPEED images randomly include a realistically rendered Earth on the background, offering a series of challenging samples to predict.
The labels for the test set of SPEED have not been disclosed meaning that any comparison done with results of SPEED will not be based on the same test set. Assuming the test set of SPEED follows the same data distribution as its train set, and in hopes of estimating the performance that LSPnet would obtain over the test set, a fixed random split of $10,000$ and $2,000$ images is performed.
All the carried trainings share the following common parameters:

\begin{enumerate}
    % \item Fixed dataset split of $10.000$ and $2.000$ samples
    \item Batch size of $N_B=16$ samples
    \item Input images are rescaled to $256\times409$ pixels (thus maintaining the aspect ratio)
    \item Adam optimizer with a learning rate of $1e-4$
    \item Learning rate decay of $1/2$ based on reaching a loss plateau
    \item Train for as many epochs as needed until convergence
    \item When HC is implemented, a total of $H=64$ non-normalized heatmaps are predicted by the Localization CNN
    \item When CDA is implemented, the hyperparameter $r$ is fixed to $0.15$
    % \item Early stopping: the best performing model over the validation split is always saved
\end{enumerate}

The loss function used to optimize the prediction of $t$ is the Mean Squared Error (MSE) loss which is presented in the following equation
\begin{equation}
    L_{position} = MSE(t, \hat{t}) = \frac{1}{N_B}\sum_{i=1}^{N_B}(t_i - \hat{t_i})^{2}
\label{translation_loss}
\end{equation}
where $t$ is the batch of ground truth translation vectors, $\hat{t}$ is the batch of predicted translation vectors, $N_B$ is the batch size and $t_i$ ($\hat{t_i}$ respectively) corresponds to the i-th translation vector within the batch.

Regarding the optimization of $(u,v)$, the loss function selected is the one proposed in the work of DSNT \cite{Dsnt} which is formulated as follows
\begin{equation}
    L_{euc}(c, \hat{c}) = ||c-\hat{c}||_2
\end{equation}
\begin{equation}
    L_{reg}(\hat{h}, \hat{c}) = D(\hat{h} || \mathcal{N}(\hat{c},\,\sigma^{2}*I_{2}))
\end{equation}
\begin{equation}
    L_{center} = \frac{1}{N_B}\sum_{i=1}^{N_B} L_{euc}(c_i, \hat{c_i}) + \lambda L_{reg}(\hat{h_i}, \hat{c_i})
\label{center_loss}
\end{equation}
where $c$ is the batch of ground truth centers $(u,v)$, $\hat{c}$ is the batch of predicted centers $(\hat{u}, \hat{v})$, $||\cdot||_2$ is the 2-norm, $\hat{h}$ is the batch of predicted normalized heatmaps, $D(\cdot||\cdot)$ is a divergence measure, $I_2$ is the $2\times2$ identity matrix, $N_B$ is the batch size and $c_i$ ($\hat{c_i}$, $\hat{h_i}$ respectively) corresponds to the i-th element within the batch. Based on experimental findings of \cite{Dsnt}, the hyperparameters $\sigma^2$ and $\lambda$ have been fixed to $1$ and the divergence measure $D(\cdot||\cdot)$ has been selected to be the Jensen-Shannon divergence. Combining both Equations (\ref{translation_loss}) and (\ref{center_loss}), the final loss for the Translation module is formed
\begin{equation}
    L_{translation} = L_{position} + L_{center}
\label{translation_module_loss}
\end{equation}
Following common practice for optimizing quaternion estimation, the following loss has been implemented
\begin{equation}
    L_{rotation} = \frac{1}{N_B}\sum_{i=1}^{N_B}2*arccos(|<q_i,\hat{q_i}>|)
\label{quaternion_module_loss}
\end{equation}
where $N_B$ is the batch size, $<\cdot, \cdot>$ represents the dot product, $|\cdot|$ is the absolute value function and $q_i$ ($\hat{q_i}$ respectively) corresponds to the i-th quaternion within the batch. Finally the complete loss for optimizing LSPnet entirely, \textit{i.e.} the Pose module, is formulated as
\begin{equation}
    L_{pose} = L_{translation} + L_{rotation}
\label{pose_module_loss}
\end{equation}

\begin{table*}[t]
\caption{LSPnet ablation study covering Orientation CNN initialization as well as HC and CDA implementation.}
\label{complete_ablation_study}
\begin{center}
\begin{tabular}{|c|c|c||c|c|}
\hline
Orientation CNN init. & HC & CDA & $E_{t}$ & $E_{q}$ ($deg$)\\
\hline
Random & \xmark & \xmark & $\pmb{0.519 \pm 1.047}$ & $36.13 \pm 41.23$ \\
\hline
ImageNet & \xmark & \xmark & $\pmb{0.519 \pm 1.047}$ & $22.36 \pm 37.33$ \\
\hline
Random & \cmark & \xmark & $0.588 \pm 1.187$ & $33.22 \pm 38.78$ \\
\hline
ImageNet & \cmark & \xmark & $0.602 \pm 1.136$ & $37.64 \pm 41.11$ \\
\hline
ImageNet & \xmark & \cmark & $\pmb{0.519 \pm 1.047}$ & $\pmb{15.70 \pm 23.61}$ \\
\hline
Random & \cmark & \cmark & $0.596 \pm 1.106$ & $33.05 \pm 36.38$ \\
\hline
\end{tabular}
\end{center}
\end{table*}

\begin{table}[ht]
\caption{Ablation study on Localization CNN enhancing the translation estimation task when connected to Position CNN.}
\label{table_loc_cnn}
\begin{center}
\begin{tabular}{|c||c|c|c|c|}
\hline
Localization CNN & $E_{x}$ & $E_{y}$ & $E_{z}$ & $E_{t}$\\
\hline
\xmark & $0.0571$ & $0.0573$ & $0.519$ & $0.539$\\
\hline
\cmark & $\pmb{0.0551}$ & $\pmb{0.0558}$ & $\pmb{0.498}$ & $\pmb{0.519}$\\
\hline
\end{tabular}
\end{center}
\end{table}

\begin{table}[ht]
\caption{Position CNN initialization study.}
\label{table_position_init}
\begin{center}
\begin{tabular}{|c||c|c|c|c|}
\hline
Initialization & $E_{x}$ & $E_{y}$ & $E_{z}$ & $E_{t}$\\
\hline
Random & $0.0746$ & $0.0816$ & $0.666$ & $0.694$\\
\hline
ImageNet & $\pmb{0.0551}$ & $\pmb{0.0558}$ & $\pmb{0.498}$ & $\pmb{0.519}$\\
\hline
\end{tabular}
\end{center}
\end{table}

A set of ablation studies have been performed in hopes of finding a highly performing LSPnet architecture specification. The following sections cover the performed ablation studies as well as a final comparison with state-of-the-art over the SPEED dataset. The error metrics used throughout the following sections are formalized here
\begin{equation}
    E_{c} = \frac{1}{N}\sum_{i=1}^{N}|t_i^c-\hat{t}_{i}^c|, \ with \ c\in\{x,y,z\}
\label{L1-coordinate-error}
\end{equation}
\begin{equation}
    E_{t} = \frac{1}{N}\sum_{i=1}^{N}|t_i-\hat{t}_{i}|_2
\label{L2-translation-error}
\end{equation}
\begin{equation}
    E_{q} = \frac{1}{N}\sum_{i=1}^{N}2*arccos(|<q_i,\hat{q_i}>|)
\label{quaternion-error}
\end{equation}
where $N$ is the size of the dataset being evaluated and $t_i^c$ ($\hat{t_i^c}$ respectively) corresponds to the coordinate value $c$ ($x$, $y$ or $z$) of the i-th translation vector within the dataset. If not specified otherwise, $E_c$ and $E_t$ are expressed in meters ($m$) while $E_{q}$ is expressed in radians ($rad$).

\subsection{Ablation studies}
During all the ablation studies performed, both CNN architectures for the Position CNN and the Orientation CNN have been fixed to ResNet18. Based on the following findings, the best performing configuration is selected and scaled up to ResNet50 before comparing to state-of-the-art.

It has been surmised that the Localization CNN aids the Position CNN in the process of translation estimation. To shed light on this claim, Table \ref{table_loc_cnn} presents translation results when using Position CNN alone in comparison to including the Localization CNN. Note that both architectures have been initialized with ImageNet weights to ensure fair comparison. A slight decrease in translation error points to the idea that Localization CNN aids, to some extent, the translation estimation task.

\begin{table*}[t]
\caption{SPEC results and comparison.}
\label{table_spec}
\begin{center}
\begin{tabular}{|c|c||c|c|c|c|}
\hline
Rank & Team & $E_{t}$ & $E_{q}$ ($deg$) & PnP\\
\hline
1 & UniAdelaide \cite{SPEC-Rank1} & $0.032\, \pm\, 0.095$ & $0.41 \pm 1.50$ & Yes\\
\hline
2 & EPFL cvlab & $0.073\, \pm\, 0.587$ & $0.91 \pm 1.29$ & Yes\\
\hline
3 & pedro fairspace \cite{SPEC-Rank3} & $0.145\, \pm\, 0.239$ & $2.49 \pm 3.02$ & No\\
\hline
- & SLAB Baseline \cite{SPEC-baseline} & $0.209\, \pm\, 1.133$ & $2.62 \pm 2.90$ & Yes\\
\hline
\vdots & \vdots & \vdots & \vdots & \vdots\\
\hline
7 & Gabrie1A & $0.318\, \pm\, 0.323$ & $12.03 \pm 12.87$ & No\\
\hline
- & LSPnet (Ours) & $0.456\, \pm\, 1.010$ & $13.96 \pm 20.13$ & No\\
\hline
8 & stainsby & $0.714\, \pm\, 1.012$ & $17.75 \pm 22.01$ & No\\
\hline
9 & VSI\_Feeney & $0.734\, \pm\, 1.273$ & $23.42 \pm 33.57$ & No\\
\hline
10 & jblumenkamp & $2.656\, \pm\, 2.149$ & $35.92 \pm 49.72$ & Yes\\
\hline
\end{tabular}
\end{center}
\end{table*}

\begin{table}[ht]
\caption{SPN and LSPnet comparison results.}
\label{table_spn}
\begin{center}
\begin{tabular}{|c||c|c|c|c|}
\hline
Model & $E_{x}$ & $E_{y}$ & $E_{z}$ & $E_{q}$ ($deg$)\\
\hline
SPN~\cite{SPEED} & $0.055$ & $0.046$ & $0.78$ & $\pmb{8.43}$\\
\hline
LSPnet (Ours) & $\pmb{0.048}$ & $\pmb{0.045}$ & $\pmb{0.44}$ & $13.96$\\
\hline
\end{tabular}
\end{center}
\end{table}

Once Localization CNN has been found to aid in optimizing the predictions of $t$, a brief comparison between ImageNet initialization and Random initialization of the Position ResNet weights has been performed and can be found in Table \ref{table_position_init}. Having fixed the initialization of the Position CNN to ImageNet weights, due to its highly positive impact, a complete ablation study has been carried covering the following architectural and training decisions: (1) Orientation CNN initialization, (2) implementation of HC and (3) implementation of CDA. Table \ref{complete_ablation_study} presents all the results obtained during this complete ablation study. Note that when HC is not implemented the Translation module used is the best one obtained among Table \ref{table_loc_cnn} and Table \ref{table_position_init}. Also note that not all the possible combinations have been tested. This is due to the findings that have been extracted throughout the ablation study process which are the following,
\begin{itemize}
    \item Similarly to Position CNN initialization, Orientation CNN ImageNet initialization significantly improves rotation estimation when not including HC, \textit{i.e.} when the input is only composed by the ROI crop.
    \item It has been found that when HC is implemented, and thus the input of the Orientation CNN is a tensor with $H + C_I$ channels, ImageNet initialization negatively impacts the orientation performance. This can be justified due to the drastically different data distributions from ImageNet with respect to the non-normalized heatmaps predicted by the Localization CNN. In this sense a random initialization is more fitting to the specialized data distribution introduced by such heatmaps.
    \item Based on the previous findings, the remaining configurations worth testing are ImageNet initialization with CDA as well as Random initialization with both HC and CDA.
    \item It is also found that when training the full Pose module at the same time, meaning that HC is implemented to enable an end-to-end differentiable training, the translation error slightly increases. This is due to the fact that in this scenario the Translation module has to be optimized to both predict the translation vector $t$ as well as transfer orientation-meaningful heatmaps to the Orientation CNN. When no HC is implemented the Translation module is completely decoupled from orientation estimation and thus it is solely trained for position-related tasks.
    \item Lastly, when randomly initializing the Orientation CNN, only implementing HC improves orientation results at the expense of slightly worsening the translation results. Furthermore, when implementing both HC and CDA the orientation results very slightly improve (and the translation error slightly increases).
\end{itemize}

Overall, the best performing configuration found does not implement HC, implements CDA and initializes the Orientation CNN with ImageNet weights. Such combination means that the translation and orientation tasks are better optimized by LSPnet when decoupled. It is worth noting how CDA greatly improves orientation estimation. %when using this configuration.

\subsection{State-of-the-art comparison}
Given all the architectural decisions taken based on the presented ablation studies, the best performing LSPnet is chosen and compared to the SPEED state-of-the-art. A first comparison is performed with the SPN network proposed in the same SPEED work \cite{SPEED}. Table \ref{table_spn} presents the obtained comparisons. LSPnet significantly surpasses SPN in depth estimation ($z$ coordinate of the translation vector). SPN, on the other hand, is capable of predicting more accurately the rotation quaternion $q$. For fair comparison, it is needed to highlight the fact that SPN relies on 3D information, performs Object detection through an off-the-shelf Object detection Deep Learning model and refines the predicted pose. Conversely, LSPnet achieves competitive results without requiring any of the aforementioned characteristics.
The final comparison is done thanks to the latest SPEC challenge \cite{SPEC} which also relied on the SPEED dataset. For this reason, the results presented in SPEC are fit to be referenced in order to localize LSPnet into the uncooperative spacecraft state-of-the-art for pose estimation. The SPEC challenge involved nearly 50 teams working towards pose estimation on the SPEED dataset for 5 months. Table \ref{table_spec} shows the obtained results by LSPnet in the context of the top 10 ranking teams of SPEC. LSPnet can be \textit{ranked} at the top 8 position. According to SPEC, a total of seven teams reconstructed the 3D model of the spacecraft to further use it in a keypoint-based solution (\textit{e.g.} in combination with a PnP solver). SPEC also found that a recurring technique across the teams is the detection of the spacecraft in the image through Object detection Deep Learning models (such as YOLO) or through image segmentation. Moreover, pose refinement steps can also be found among the teams (\textit{e.g.} \cite{SPEC-Rank1}). LSPnet is capable of \textit{ranking} top 8 while not relying on any 3D information, not refining the pose and implementing a simple yet efficient Object detection technique (easily augmented by CDA). Note that even though we are directly comparing LSPnet results with SPEED state-of-the-art results, our results are based on a different test set. This means that all the comparisons done through this section should be taken as indications of how LSPnet performs with respect to the state-of-the-art.
% Some qualitative insights are presented to conclude the experiments section. The top 3 successful spacecraft's center predictions are offered in Figure \ref{fig:long-successful-predictions} (for long range scenarios) and in Figure \ref{fig:short-successful-predictions} (for short range scenarios) as well as the top 3 failure predictions in Figure \ref{fig:failure-predictions}. Notice how the failure cases correspond to challenging images in which the spacecraft can easily be confused with its background partly due to the large distance from the camera to the spacecraft.
%-------------------------------------------------------------------------
\section{Conclusions}
The goal of the here presented work is to provide the space literature with a simpler yet still effective solution for pose estimation of uncooperative spacecrafts which does not require prior 3D information nor involves pose refinement. Additionally, the proposed model is capable of generating bounding boxes without relying on a complex Object detection model and without needing bounding boxes labels (only requiring translation ground truths). It is shown how LSPnet achieves comparable results with respect to SPEED state-of-the-art. Extensions of this work may target spacecraft generalization as well as pose tracking.

{\small
\bibliographystyle{ieee_fullname}
\bibliography{egbib}
}

\end{document}